\documentclass{article}

\usepackage{PRIMEarxiv}
\usepackage[utf8]{inputenc}
\usepackage[T1]{fontenc}
\usepackage{hyperref}
\usepackage{url}
\usepackage{booktabs}
\usepackage{amsfonts}
\usepackage{nicefrac}
\usepackage{microtype}
\usepackage{fancyhdr}
\usepackage{graphicx}
\usepackage{amsmath}
\usepackage[table]{xcolor}
\usepackage{float}
\usepackage{placeins}
\usepackage{tikz}
\usepackage{pgfplots}
\usepackage{etoolbox}

\usetikzlibrary{arrows.meta,backgrounds,calc,fit,positioning,shapes.geometric}
\pgfplotsset{compat=1.18}

\graphicspath{{figures/}}
\definecolor{paceNavy}{HTML}{173B63}
\definecolor{paceInk}{HTML}{223044}
\definecolor{paceGold}{HTML}{E9C46A}
\definecolor{paceTeal}{HTML}{2A9D8F}
\definecolor{paceOrange}{HTML}{F4A261}
\definecolor{paceRed}{HTML}{D1495B}
\definecolor{paceGreen}{HTML}{729E4D}
\definecolor{paceBlueLight}{HTML}{DCE7F5}
\definecolor{paceTealLight}{HTML}{D8EEE8}
\definecolor{paceOrangeLight}{HTML}{F6DDD0}
\definecolor{paceGreenLight}{HTML}{DCEAD5}
\definecolor{paceGray}{HTML}{D5DADF}
\makeatletter
\patchcmd{\thebibliography}{\small}{\normalsize\raggedright}{}{}
\patchcmd{\thebibliography}{\sloppy\clubpenalty4000\widowpenalty4000}{\sloppy\clubpenalty4000\widowpenalty4000\itemsep 2pt}{}{}
\makeatother
\newcommand{\tighttable}[1]{%
  \begingroup
  \setlength{\tabcolsep}{5pt}%
  \renewcommand{\arraystretch}{1.12}%
  #1%
  \endgroup
}

\title{Pangu-ACE: Adaptive Cascaded Experts for Educational Response Generation on EduBench}

\author{
Dinghao Li, Wenlong Zhou, Zhimin Chen, Yuehan Peng, Hong Ni, Chengfu Zou, Guoyu Shi, Yaochen Li\thanks{Corresponding author: yaochenli@mail.xjtu.edu.cn} 
}

\begin{document}
\maketitle

\begin{abstract}
Educational assistants should spend more computation only when the task needs it. This paper rewrites our earlier draft around the system that was actually implemented and archived in the repository: a sample-level \texttt{1B} $\rightarrow$ \texttt{7B} cascade for the shared-8 EduBench benchmark. The final system, \textbf{Pangu-ACE}, uses a \texttt{1B} tutor-router to produce a draft answer plus routing signals, then either accepts the draft or escalates the sample to a \texttt{7B} specialist prompt. We also correct a major offline evaluation bug: earlier summaries over-credited some open-form outputs that only satisfied superficial format checks. After CPU-side rescoring from saved prediction JSONL, the full Chinese test archive (\(7013\) samples) shows that \texttt{cascade\_final} improves deterministic quality from \(0.457\) to \(0.538\) and format validity from \(0.707\) to \(0.866\) over the legacy \texttt{rule\_v2} system while accepting \(19.7\%\) of requests directly at \texttt{1B}. Routing is strongly task dependent: \texttt{IP} is accepted by \texttt{1B} \(78.0\%\) of the time, while \texttt{QG} and \texttt{EC} still escalate almost always. The current archived deployment does not yet show latency gains, so the defensible efficiency story is routing selectivity rather than wall-clock speedup. We also package a reproducible artifact-first paper workflow and clarify the remaining external-baseline gap: GPT-5.4 re-judging is implemented locally, but the configured provider endpoint and key are invalid, so final sampled-baseline alignment with GPT-5.4 remains pending infrastructure repair.
\end{abstract}


\section{Introduction}
Educational use cases are one of the clearest settings in which language models must balance speed, structure, and pedagogical quality \cite{xu2024large, agent_survey, edubench}. A short factual question, a grading rubric, a personalized study plan, and a teaching-material outline should not all consume the same inference budget, nor should they be judged with the same output contract. That heterogeneity makes education a natural setting for sample-level routing: the system should spend more computation only when the task demands it.

The repository underlying this paper is not a generic agent demo. It is a paper-first offline benchmark stack with deterministic shared-8 EduBench splits, traceable per-sample predictions, and CPU-side recomputation from saved artifacts. The implemented system centers on a calibrated \texttt{1B} $\rightarrow$ \texttt{7B} cascade rather than the quantization-heavy design described in an earlier draft. A lightweight \texttt{1B} stage produces a draft answer together with routing signals; a \texttt{7B} specialist prompt is invoked only for higher-risk samples. This design is closest to cost-aware routing and cascade methods such as FrugalGPT and RouteLLM \cite{frugalgpt, routellm}, while also borrowing from the broader agent literature on staged reasoning and action, including ReAct and Reflexion \cite{react, reflexion}.

The system is also motivated by dual-process reasoning: cheap drafts should handle routine cases, while more deliberate specialist computation should be reserved for harder ones \cite{kahneman2011thinking, system2attention, quiet_star}. Likewise, the drafting and escalation logic is downstream from the chain-of-thought literature \cite{wei2022chain, self_consistency}, but the unit of control in our setting is not token-by-token reasoning length. It is the full educational sample.

Rewriting the paper around the actual code and artifacts reveals a second issue: the original offline summaries overstated quality on open-form tasks. The bug was simple but consequential. Outputs that looked syntactically valid, including router scaffolding such as literal \texttt{draft\_answer} placeholders, were sometimes counted as successful final responses. We therefore rebuilt the paper around rescored predictions, stricter parsing, and explicit limitations.

This revision makes four contributions:
\begin{enumerate}
    \item We present \textbf{Pangu-ACE}, the implemented adaptive cascade used in the final EduBench experiments, with explicit task-family routing and specialist refinement.
    \item We define the paper's benchmark protocol around the shared-8 EduBench tasks and deterministic saved-artifact analysis rather than ad hoc online evaluation.
    \item We correct the offline metric bug for open-form tasks and recompute the paper's quantitative results from raw prediction files.
    \item We diagnose the missing GPT-5.4 baseline path as a provider-configuration failure, not a scoring-pipeline failure, and provide a reproducible re-judge workflow once valid credentials exist.
\end{enumerate}

\section{Related Work}
\subsection{Educational Agents and Benchmarks}
Large language models are increasingly framed as educational agents rather than generic chatbots, with responsibilities that include explanation, grading, personalization, and instructional planning \cite{xu2024large, agent_survey}. EduBench makes this framing concrete by evaluating a broad set of educational scenarios rather than a single-answer benchmark \cite{edubench}. That benchmark design is important for our setting because it exposes task heterogeneity directly: the same system must handle structured grading, pedagogical hints, content generation, and planning-like outputs.

\subsection{Cost-Aware Routing and Cascades}
Our system is closest to the line of work that routes requests across models or execution paths to trade cost against quality. FrugalGPT studies cascaded model selection under budget constraints \cite{frugalgpt}, while RouteLLM focuses on routing prompts between strong and weak LLMs \cite{routellm}. Pangu-ACE follows the same high-level philosophy, but adapts it to a paper-first offline educational setting with deterministic splits, structured output contracts, and task-family specialist prompts. The resulting unit of routing is the full educational sample, not a generic query with a scalar preference score.

\subsection{Inference Efficiency and Serving}
Routing policy is only one component of efficient language-model systems. Kernel- and cache-level methods such as FlashAttention, FlashAttention-2, and PagedAttention reduce the cost of each model call \cite{flashattention, flashattention2, pagedattention}, while speculative and multi-head decoding methods such as speculative decoding, Medusa, and EAGLE target generation-time overhead directly \cite{speculative_decoding, medusa, eagle}. We cite this line of work to sharpen our scope: Pangu-ACE studies \emph{which} model path an educational sample should follow, whereas the serving literature studies \emph{how} each invocation is made faster once the path is chosen \cite{pangu_embedded}.

\subsection{Reasoning, Reflection, and Structured Improvement}
The \texttt{1B} draft followed by selective \texttt{7B} refinement is also related to the broader literature on explicit reasoning and iterative improvement. Chain-of-thought prompting and self-consistency show that multi-step reasoning can improve solution quality at the cost of longer generation \cite{wei2022chain, self_consistency}. ReAct and Toolformer demonstrate how reasoning can be staged around structured decisions and intermediate actions \cite{react, toolformer}, while Reflexion explores self-improvement through intermediate critique \cite{reflexion}. Our system does not claim a new reasoning paradigm. Instead, it packages these ideas into a deterministic educational cascade in which structured drafting, risk scoring, and specialist repair are all visible in saved artifacts.

\section{Benchmark and System}
\subsection{Shared-8 EduBench Setup}
We use the shared-8 EduBench task set \cite{edubench}: \texttt{Q\&A}, \texttt{AG}, \texttt{EC}, \texttt{IP}, \texttt{PCC}, \texttt{PLS}, \texttt{QG}, and \texttt{TMG}. The benchmark excludes \texttt{ES} from the main paper because it is not present symmetrically across the archived Chinese and English assets. The deterministic split plan in the repository fixes the Chinese shared-8 total at \(7793\) samples and the English shared-8 total at \(8102\), with seed \(42\) stratification and frozen dev/test assignments.

The main archived result used in this paper is the full Chinese test split of \(7013\) samples. Only \texttt{cascade\_final} and \texttt{rule\_v2} were fully preserved for that split, so the main table isolates whether the new routing stack improves over the legacy system on exactly the same artifact pool. To retain context for \texttt{1B}-only and \texttt{7B}-only behavior, we also use the preserved fast diagnostic subsets: \(354\) Chinese test samples and \(368\) English test samples.

\subsection{Pangu-ACE Pipeline}
\begin{figure}[H]
  \centering
  \resizebox{0.985\linewidth}{!}{%
\begin{tikzpicture}[
  >=Latex,
  font=\small,
  stage/.style={
    rounded corners=8pt,
    draw=paceInk,
    very thick,
    align=center,
    fill=white,
    minimum width=3.05cm,
    minimum height=1.95cm,
    text width=2.6cm,
    inner sep=7pt
  },
  prep/.style={
    rounded corners=8pt,
    draw=paceInk,
    thick,
    align=center,
    fill=paceGray!28,
    minimum width=2.8cm,
    minimum height=0.95cm,
    text width=2.35cm,
    inner sep=6pt
  },
  branch/.style={
    rounded corners=8pt,
    draw=paceInk,
    very thick,
    align=center,
    minimum width=3.35cm,
    minimum height=1.9cm,
    text width=2.8cm,
    inner sep=8pt
  },
  panel/.style={draw=paceInk!70, rounded corners=10pt, dashed, inner sep=10pt},
  badge/.style={
    circle,
    draw=paceInk,
    line width=0.9pt,
    fill=white,
    minimum size=0.55cm,
    inner sep=0pt,
    font=\scriptsize\bfseries
  },
  chip/.style={
    rounded corners=5pt,
    draw=paceInk!55,
    fill=white,
    minimum height=0.42cm,
    inner sep=2.5pt,
    font=\scriptsize\bfseries
  },
  flow/.style={-Latex, very thick, draw=paceNavy},
  support/.style={-Latex, thick, draw=paceInk!65},
  note/.style={font=\scriptsize, text=paceInk!82}
]

% OFFLINE PREP
\node[prep] (splits) at (3.8,2.8) {\textbf{Deterministic splits}\\shared-8 benchmark};
\node[prep] (prompts) at (7.6,2.8) {\textbf{Prompt bank}\\family specialists};
\node[prep] (freeze) at (11.4,2.8) {\textbf{Threshold freeze}\\dev calibration};

% MAIN PIPELINE
\node[stage, fill=paceGold!80] (input) at (0,0)
  {\textbf{EduBench sample}\\prompt, task, language};

\node[stage] (normalize) at (3.8,0)
  {\textbf{Request\\normalizer}\\task key, family, output contract};

\node[stage, fill=paceTealLight] (router) at (7.6,0)
  {\textbf{1B tutor-router}\\draft answer + route cues};

% CUSTOM RISK CALIBRATOR
\node[stage, fill=paceBlueLight, minimum height=2.1cm] (risk) at (11.4,0) {};
\node[align=center, font=\small] at ($(risk.north)+(0,-0.6)$) {\textbf{Risk calibrator}\\frozen dev thresholds};
\begin{scope}[shift={(risk.south west)}]
  \draw[rounded corners=3pt, draw=paceInk!45, fill=white] (0.4,0.3) rectangle (2.65,0.7);
  \fill[paceGreen]  (0.5,0.36) rectangle (0.9,0.64);
  \fill[paceTeal]   (1.0,0.36) rectangle (1.4,0.58);
  \fill[paceOrange] (1.5,0.36) rectangle (1.9,0.64);
  \fill[paceRed]    (2.0,0.36) rectangle (2.55,0.85);
  \draw[paceRed, thick] (2.35,0.32) -- (2.35,0.89);
\end{scope}

% CUSTOM ROUTING PROFILE
\node[stage, minimum width=3.5cm, minimum height=2.3cm] (profile) at (15.2,0) {};
\node[align=center, font=\small] at ($(profile.north)+(0,-0.6)$) {\textbf{Routing profile}\\task-dependent escalation};
\begin{scope}[shift={($(profile.south)+(-1.6,0.2)$)}]
  \node[anchor=west, font=\scriptsize] at (0.4, 0.95) {IP};
  \fill[paceGreen] (1.0, 0.85) rectangle (2.1, 1.05);
  \fill[paceNavy]  (2.1, 0.85) rectangle (2.6, 1.05);

  \node[anchor=west, font=\scriptsize] at (0.4, 0.55) {PLS};
  \fill[paceGreen] (1.0, 0.45) rectangle (1.6, 0.65);
  \fill[paceNavy]  (1.6, 0.45) rectangle (2.6, 0.65);

  \node[anchor=west, font=\scriptsize] at (0.4, 0.15) {QG};
  \draw[paceGreen, thick] (1.0, 0.0) -- (1.0, 0.25);
  \fill[paceNavy]  (1.0, 0.05) rectangle (2.6, 0.25);
\end{scope}

% BRANCHES
\node[branch, fill=paceGreenLight] (accept) at (5.7,-4.0)
  {\textbf{Direct answer path}\\accept validated 1B output};

\node[branch, fill=paceOrangeLight, minimum width=3.6cm, text width=3.0cm] (specialist) at (11.4,-4.0)
  {\textbf{7B specialist repair}\\reasoning, assessment, planning};

% VALIDATION & ARTIFACTS
\node[stage, minimum width=4.65cm, minimum height=1.7cm, text width=4.0cm] (validator) at (8.55,-7.0)
  {\textbf{Validation and repair}\\reject scaffolds, repair malformed fields, finalize answer};

\node[stage, fill=paceGray!55, minimum width=6.5cm, minimum height=1.25cm, text width=5.9cm] (artifacts) at (8.55,-9.0)
  {\textbf{Artifacts and audit trail}\\predictions \;|\; traces \;|\; summaries \;|\; judge cache};

% BADGES
\node[badge, fill=paceGold!45] at (input.north west) {E};
\node[badge, fill=paceBlueLight] at (normalize.north west) {N};
\node[badge, fill=paceTealLight!75] at (router.north west) {1B};
\node[badge, fill=paceBlueLight!70] at (risk.north west) {R};
\node[badge, fill=paceOrangeLight!75] at (profile.north west) {T};
\node[badge, fill=paceGreenLight!75] at (accept.north west) {A};
\node[badge, fill=paceOrangeLight!85] at (specialist.north west) {7B};
\node[badge, fill=paceGray!18] at (validator.north west) {V};
\node[badge, fill=paceGray!35] at (artifacts.north west) {L};

% BACKGROUND ROUTING
\begin{scope}[on background layer]
  \node[panel, fit=(input)(normalize)(router)(risk)(profile)(accept)(specialist)(validator)(artifacts)] (onlinepanel) {};
  \node[panel, fit=(splits)(prompts)(freeze)] (offlinepanel) {};

  \draw[flow] (input) -- (normalize);
  \draw[flow] (normalize) -- (router);
  \draw[flow] (router) -- (risk);
  \draw[flow] (risk) -- (profile);

  \draw[flow] (risk.south) -- (specialist.north) node[midway, right, font=\scriptsize\bfseries, text=paceRed, xshift=2pt] {high risk};
  \draw[flow, rounded corners=6pt] ($(risk.south)+(0,-0.8)$) -| (accept.north) node[pos=0.25, above, font=\scriptsize\bfseries, text=paceGreen!60!black] {low risk};

  \draw[flow, rounded corners=6pt] (accept.south) -- +(0,-0.8) -| (validator.north);
  \draw[flow, rounded corners=6pt] (specialist.south) -- +(0,-0.8) -| (validator.north);

  \draw[flow] (validator) -- (artifacts);

  \draw[support] (splits) -- (normalize);
  \draw[support] (freeze) -- (risk);
  \draw[support, rounded corners=6pt] (prompts.south) -- +(0,-0.4) -| (9.5, -2.0) |- (specialist.west);
\end{scope}

% CHIPS
\node[chip] at (input.south) {Q\&A \quad AG \quad EC \quad QG};
\node[chip] at (normalize.south) {ZH / EN \quad JSON schema};
\node[chip] at (router.south) {draft \quad family \quad confidence};
\node[chip] at (accept.south) {fast \quad low-cost \quad valid};
\node[chip] at (specialist.south) {reason \quad grade \quad plan};
\node[chip] at (validator.south) {schema \quad substance \quad trace};

% MARGINALIA
\node[anchor=south west, font=\scriptsize\bfseries, text=paceNavy]
  at ($(onlinepanel.north west)+(0.02,0.08)$) {Online inference path};
\node[anchor=south west, font=\scriptsize\bfseries, text=paceNavy]
  at ($(offlinepanel.north west)+(0.02,0.08)$) {Offline preparation};

\node[note, anchor=east, align=right] at ($(accept.west)+(-0.3, 0)$)
  {Routine requests\\stop at 1B.};
\node[note, anchor=west, align=left] at ($(specialist.east)+(0.3, 0)$)
  {Harder samples pay\\for 7B repair\\only when needed.};
\end{tikzpicture}%
}
  \caption{Implemented Pangu-ACE pipeline. A normalized EduBench sample is drafted by the \texttt{1B} tutor-router, scored by the calibrator, then either accepted directly or sent to a \texttt{7B} specialist before final validation and artifact logging. The inset emphasizes that routing behavior is task dependent rather than uniform.}
  \label{fig:system_overview}
\end{figure}
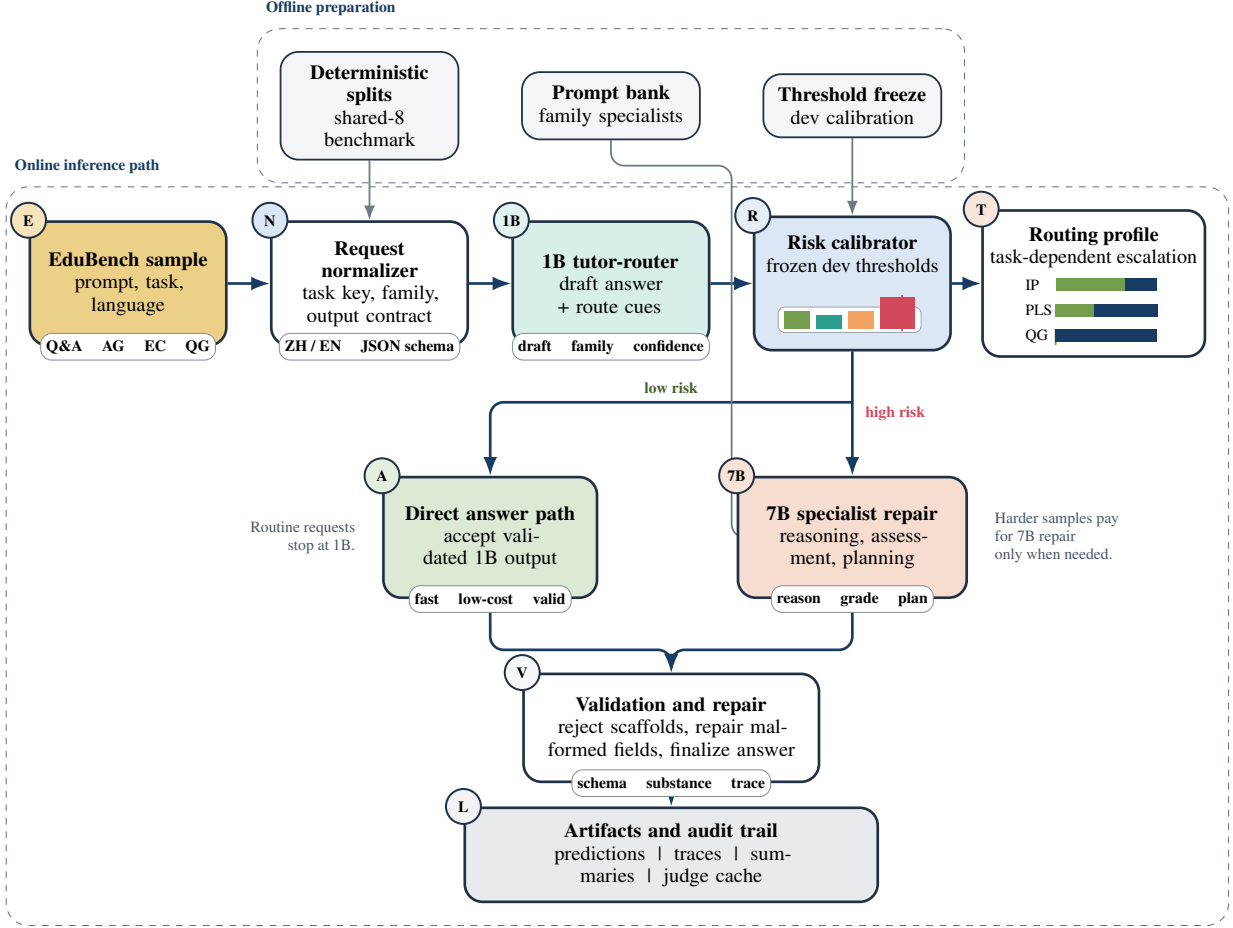

Figure \ref{fig:system_overview} shows the final pipeline. The core design is sample-level semantic routing, not token-level speculative decoding:
\begin{enumerate}
    \item \textbf{Request normalization}: map each EduBench sample to a shared schema with task key, task family, language, expected output format, and prompt text.
    \item \textbf{1B tutor-router}: generate a low-cost draft answer plus routing metadata such as task-family prediction, confidence, and format signals.
    \item \textbf{Risk calibration}: compare the sample's risk score against frozen thresholds obtained on the dev split.
    \item \textbf{Specialist refinement}: if the sample is high risk, pass the original request together with the \texttt{1B} draft to a \texttt{7B} specialist prompt organized by task family (reasoning, assessment, planning).
    \item \textbf{Validation and repair}: normalize the final output, reject router scaffolds and placeholders, and save all artifacts for later CPU-side analysis.
\end{enumerate}

\FloatBarrier

The system's strength is not universal. It is best viewed as a calibrated routing layer on top of specialist prompting. Strong planning and generation tasks still depend heavily on \texttt{7B}, while idea prompting and parts of personalized learning are often handled successfully by \texttt{1B}.

\section{Evaluation Protocol and Metric Correction}
\subsection{Artifact-First Evaluation}
The paper follows the repository's artifact-first protocol: run inference once, save prediction JSONL and trace JSONL, then regenerate summaries later without re-running GPUs. This matters because the current paper is built after the experimental runs were finished. The rewritten tables and figures are produced from saved artifacts in \path{outputs/predictions/} and \path{outputs/traces/}, not from manually copied console logs.

\subsection{Corrected Deterministic Quality}
Closed-form tasks and open-form tasks need different deterministic checks. For \texttt{Q\&A}, \texttt{AG}, and \texttt{EC}, we use exact or field-level comparison against the ground truth. For open-form tasks, we use a stricter criterion than the earlier draft:
\[
q_i =
\begin{cases}
1[\hat{y}_i = y_i], & \text{for } \texttt{Q\&A}, \texttt{AG}, \texttt{EC} \\
1[\mathrm{valid}(\hat{y}_i) \land \mathrm{substantive}(\hat{y}_i)], & \text{for } \texttt{IP}, \texttt{PCC}, \texttt{PLS}, \texttt{QG}, \texttt{TMG}
\end{cases}
\]
where \(\mathrm{substantive}(\hat{y}_i)\) rejects placeholders, router scaffolding, and malformed nested drafts. This correction matters because superficially valid JSON should not count as a successful educational response when its content is only a scaffold.

We report four families of metrics:
\begin{itemize}
    \item \textbf{Quality}: the deterministic score above, averaged over the relevant samples.
    \item \textbf{Format validity}: whether the output satisfies the expected schema after normalization.
    \item \textbf{Routing rates}: \texttt{1B} acceptance and \texttt{7B} invocation extracted from saved traces.
    \item \textbf{Latency}: archived end-to-end latency per sample. This reflects the detached service setup that was actually used during the runs.
\end{itemize}

\section{Results}
\subsection{Main Archived Chinese Test}
Table \ref{tab:main_zh} is the paper's primary quantitative result. On the same \(7013\)-sample Chinese test archive, \texttt{cascade\_final} raises quality by \(+0.081\) absolute and format validity by \(+0.159\) absolute over \texttt{rule\_v2}. It also accepts \(19.7\%\) of requests directly at \texttt{1B}, reducing the fraction that reaches the \texttt{7B} stage from \(100\%\) to \(80.3\%\).

The trade-off is equally important: end-to-end latency is \emph{higher} for the cascade in the current deployment (\(26.35\)s vs. \(18.33\)s). The paper therefore should not claim a latency win. The defensible claim is narrower and more useful: the new system buys better quality and cleaner outputs while demonstrating selective \texttt{7B} usage. Recovering wall-clock gains will require implementation work beyond routing itself, such as fusing the router stage, reducing orchestration overhead, or shortening specialist outputs.

This gap between selective routing and realized speedup is consistent with the serving literature: once an extra routing stage is introduced, end-to-end latency can remain dominated by kernel efficiency, KV-cache management, and decoding overhead rather than by model choice alone \cite{flashattention, pagedattention, speculative_decoding, medusa, eagle}.

\begin{table}[H]
\centering
\caption{Full Chinese test results after rescoring saved predictions.}
\label{tab:main_zh}
\small
\tighttable{\begin{tabular}{lrrrrrr}
\toprule
System & Samples & Quality & Format & Latency (s) & 7B rate & 1B accept \\
\midrule
\texttt{rule\_v2} & 7013 & 0.457 & 0.707 & 18.33 & 100.0\% & -- \\
\rowcolor{paceTealLight}
\texttt{cascade\_final} & 7013 & \textbf{0.538} & \textbf{0.866} & 26.35 & \textbf{80.3\%} & \textbf{19.7\%} \\
\bottomrule
\end{tabular}
}
\end{table}
\FloatBarrier

\subsection{Fast Diagnostic Baselines}
The fast diagnostic subsets provide the missing baselines that were not archived on the full Chinese test. On the Chinese subset, \texttt{7b\_only} reaches quality \(0.591\) and \texttt{cascade\_final} reaches \(0.510\) while invoking \texttt{7B} on \(77.1\%\) of samples. On the English subset, \texttt{7b\_only} reaches \(0.534\) and \texttt{cascade\_final} reaches \(0.472\) with \texttt{7B} invocation at \(85.3\%\). \texttt{1b\_only} is consistently much weaker in both languages.

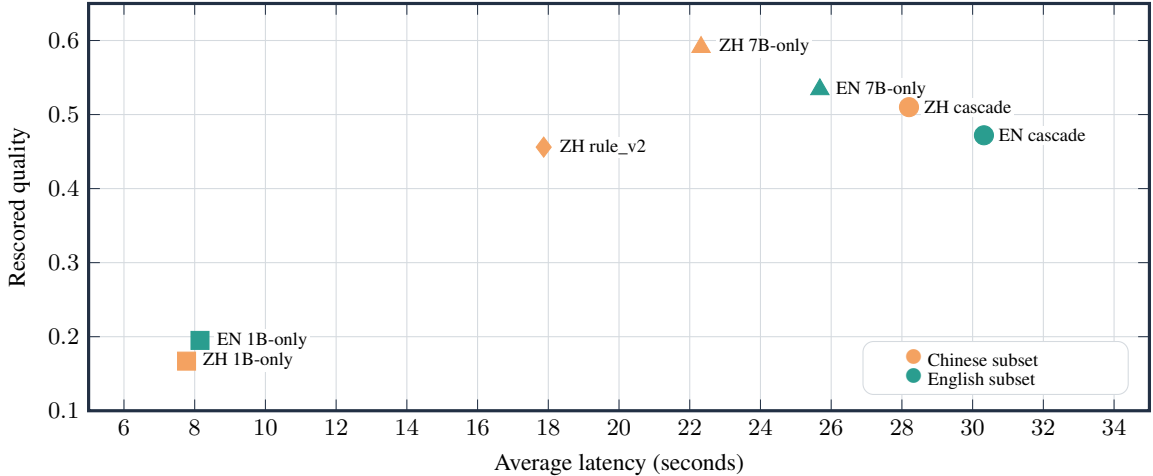
\begin{figure}[H]
  \centering
  \begin{tikzpicture}
\begin{axis}[
  width=0.95\linewidth,
  height=7.0cm,
  xmin=5,
  xmax=35,
  ymin=0.10,
  ymax=0.65,
  xlabel={Average latency (seconds)},
  ylabel={Rescored quality},
  axis line style={draw=paceInk, very thick},
  tick style={draw=paceInk},
  tick label style={font=\small},
  label style={font=\small},
  grid=major,
  major grid style={draw=paceGray},
  clip=false
]
\addplot[only marks, mark=square*, mark size=3.4pt, color=paceOrange] coordinates {(7.77,0.167)};
\node[anchor=west, font=\scriptsize, fill=white, inner sep=1pt] at (axis cs:8.15,0.167) {ZH 1B-only};

\addplot[only marks, mark=diamond*, mark size=3.8pt, color=paceOrange] coordinates {(17.87,0.456)};
\node[anchor=west, font=\scriptsize, fill=white, inner sep=1pt] at (axis cs:18.25,0.456) {ZH rule\_v2};

\addplot[only marks, mark=*, mark size=3.6pt, color=paceOrange] coordinates {(28.20,0.510)};
\node[anchor=west, font=\scriptsize, fill=white, inner sep=1pt] at (axis cs:28.55,0.510) {ZH cascade};

\addplot[only marks, mark=triangle*, mark size=4.0pt, color=paceOrange] coordinates {(22.32,0.591)};
\node[anchor=west, font=\scriptsize, fill=white, inner sep=1pt] at (axis cs:22.75,0.591) {ZH 7B-only};

\addplot[only marks, mark=square*, mark size=3.4pt, color=paceTeal] coordinates {(8.15,0.195)};
\node[anchor=west, font=\scriptsize, fill=white, inner sep=1pt] at (axis cs:8.55,0.195) {EN 1B-only};

\addplot[only marks, mark=*, mark size=3.6pt, color=paceTeal] coordinates {(30.32,0.472)};
\node[anchor=west, font=\scriptsize, fill=white, inner sep=1pt] at (axis cs:30.65,0.472) {EN cascade};

\addplot[only marks, mark=triangle*, mark size=4.0pt, color=paceTeal] coordinates {(25.68,0.534)};
\node[anchor=west, font=\scriptsize, fill=white, inner sep=1pt] at (axis cs:26.05,0.534) {EN 7B-only};

\draw[draw=paceGray, rounded corners=4pt, fill=white] (rel axis cs:0.73,0.04) rectangle (rel axis cs:0.98,0.17);
\node[anchor=west, font=\scriptsize] at (rel axis cs:0.76,0.13) {\textcolor{paceOrange}{\Large$\bullet$} Chinese subset};
\node[anchor=west, font=\scriptsize] at (rel axis cs:0.76,0.08) {\textcolor{paceTeal}{\Large$\bullet$} English subset};
\end{axis}
\end{tikzpicture}
  \caption{Quality-latency trade-off on the preserved fast diagnostic subsets. The present cascade improves strongly over \texttt{1b\_only}, but it still trails \texttt{7b\_only} in wall-clock latency because routing overhead is not yet amortized.}
  \label{fig:quality_latency}
\end{figure}

Figure \ref{fig:quality_latency} summarizes the resulting quality-latency frontier. The current archived cascade is not faster than \texttt{7b\_only}; in fact it is slower because the extra routing stage adds orchestration cost. The value of the figure is therefore diagnostic rather than celebratory: it shows that the routing stack is not yet a free lunch. We gain selectivity, but we have not yet converted that selectivity into latency savings.

\FloatBarrier

\subsection{Routing Behavior by Task}
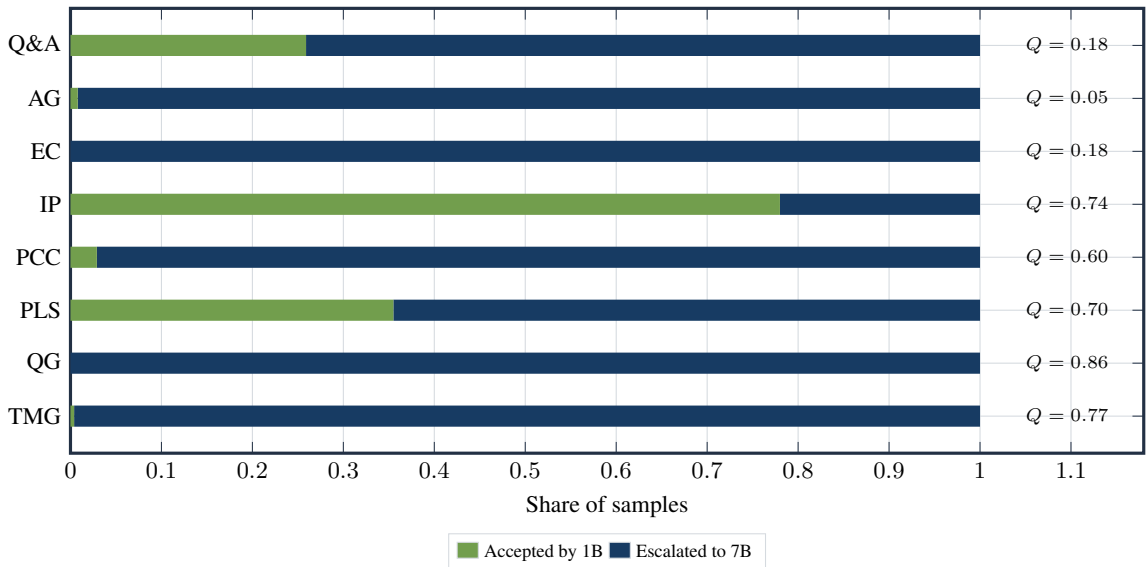
\begin{figure}[H]
  \centering
  \begin{tikzpicture}
\begin{axis}[
  width=0.96\linewidth,
  height=7.5cm,
  xbar stacked,
  xmin=0,
  xmax=1.18,
  symbolic y coords={Q\&A,AG,EC,IP,PCC,PLS,QG,TMG},
  ytick=data,
  y dir=reverse,
  bar width=8pt,
  xlabel={Share of samples},
  axis line style={draw=paceInk, very thick},
  tick style={draw=paceInk},
  tick label style={font=\small},
  label style={font=\small},
  grid=major,
  major grid style={draw=paceGray},
  legend style={draw=paceGray, fill=white, font=\scriptsize, at={(0.5,-0.18)}, anchor=north, legend columns=2},
  clip=false
]
\addplot[fill=paceGreen, draw=none] coordinates {
  (0.259,Q\&A)
  (0.008,AG)
  (0.000,EC)
  (0.780,IP)
  (0.029,PCC)
  (0.355,PLS)
  (0.000,QG)
  (0.004,TMG)
};
\addplot[fill=paceNavy, draw=none] coordinates {
  (0.741,Q\&A)
  (0.992,AG)
  (1.000,EC)
  (0.220,IP)
  (0.971,PCC)
  (0.645,PLS)
  (1.000,QG)
  (0.996,TMG)
};
\legend{Accepted by 1B,Escalated to 7B}

\node[anchor=west, font=\scriptsize] at (axis cs:1.04,Q\&A) {$Q=0.18$};
\node[anchor=west, font=\scriptsize] at (axis cs:1.04,AG) {$Q=0.05$};
\node[anchor=west, font=\scriptsize] at (axis cs:1.04,EC) {$Q=0.18$};
\node[anchor=west, font=\scriptsize] at (axis cs:1.04,IP) {$Q=0.74$};
\node[anchor=west, font=\scriptsize] at (axis cs:1.04,PCC) {$Q=0.60$};
\node[anchor=west, font=\scriptsize] at (axis cs:1.04,PLS) {$Q=0.70$};
\node[anchor=west, font=\scriptsize] at (axis cs:1.04,QG) {$Q=0.86$};
\node[anchor=west, font=\scriptsize] at (axis cs:1.04,TMG) {$Q=0.77$};
\end{axis}
\end{tikzpicture}
  \caption{Routing mix on the full Chinese test archive. Bars show how often each task is accepted at \texttt{1B} or escalated to \texttt{7B}; the annotation on the right reports rescored quality for the same task.}
  \label{fig:routing_by_task}
\end{figure}

The routing analysis in Figure \ref{fig:routing_by_task} explains why the cascade is still worthwhile. The system is not uniformly conservative. It accepts \texttt{IP} samples directly at \texttt{1B} \(78.0\%\) of the time, \texttt{PLS} \(35.5\%\), and \texttt{Q\&A} \(25.9\%\). In contrast, \texttt{EC} and \texttt{QG} escalate every sample, and \texttt{TMG} escalates \(99.6\%\) of them. This is the right qualitative pattern for the current prompt inventory: ideation is relatively draft-friendly, while generation and correction tasks remain specialist-heavy.

\FloatBarrier

\subsection{Task-Level Gains and Failure Modes}
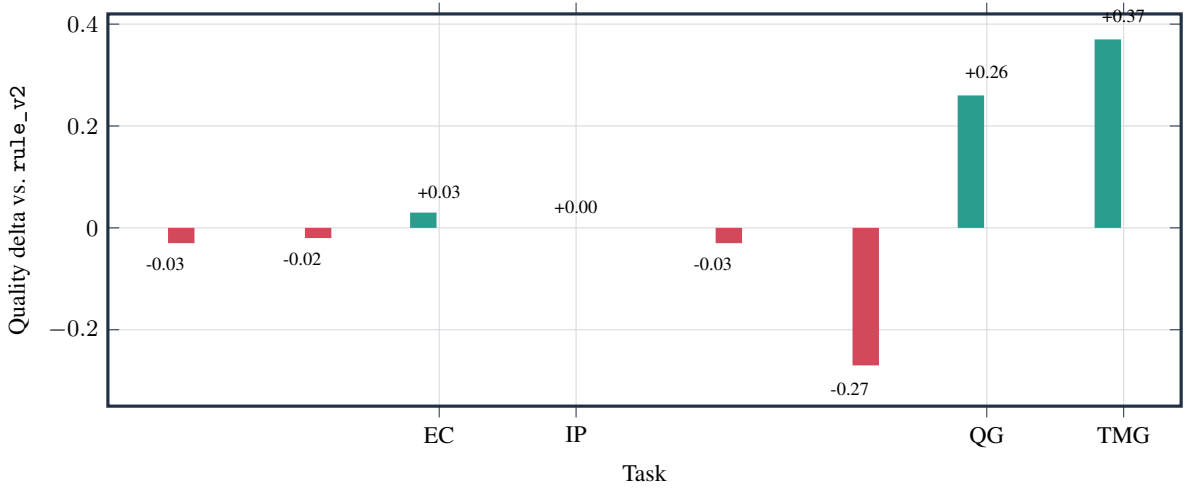
\begin{figure}[H]
  \centering
  \begin{tikzpicture}
\begin{axis}[
  width=0.96\linewidth,
  height=6.8cm,
  ybar,
  ymin=-0.35,
  ymax=0.42,
  symbolic x coords={Q\&A,AG,EC,IP,PCC,PLS,QG,TMG},
  xtick=data,
  xlabel={Task},
  ylabel={Quality delta vs.\ \texttt{rule\_v2}},
  axis line style={draw=paceInk, very thick},
  tick style={draw=paceInk},
  tick label style={font=\small},
  x tick label style={rotate=0, anchor=north},
  label style={font=\small},
  grid=major,
  major grid style={draw=paceGray},
  enlarge x limits=0.06,
  bar width=10pt,
  clip=false
]
\addplot[fill=paceTeal, draw=none] coordinates {
  (EC,0.03)
  (IP,0.00)
  (QG,0.26)
  (TMG,0.37)
};
\addplot[fill=paceRed, draw=none] coordinates {
  (Q\&A,-0.03)
  (AG,-0.02)
  (PCC,-0.03)
  (PLS,-0.27)
};

\node[font=\scriptsize, anchor=south] at (axis cs:EC,0.04) {+0.03};
\node[font=\scriptsize, anchor=south] at (axis cs:IP,0.01) {+0.00};
\node[font=\scriptsize, anchor=south] at (axis cs:QG,0.275) {+0.26};
\node[font=\scriptsize, anchor=south] at (axis cs:TMG,0.385) {+0.37};

\node[font=\scriptsize, anchor=north] at (axis cs:Q\&A,-0.04) {-0.03};
\node[font=\scriptsize, anchor=north] at (axis cs:AG,-0.03) {-0.02};
\node[font=\scriptsize, anchor=north] at (axis cs:PCC,-0.04) {-0.03};
\node[font=\scriptsize, anchor=north] at (axis cs:PLS,-0.285) {-0.27};
\end{axis}
\end{tikzpicture}
  \caption{Per-task quality delta of \texttt{cascade\_final} relative to \texttt{rule\_v2} on the full Chinese test archive. The largest gains are on question generation and teaching-material generation; grading and personalized learning remain weak points.}
  \label{fig:task_delta}
\end{figure}

Figure \ref{fig:task_delta} compares per-task quality against the legacy \texttt{rule\_v2} baseline on the full Chinese archive. The cascade helps most on generation-heavy tasks: \texttt{QG} improves by \(+0.265\) and \texttt{TMG} by \(+0.366\). It is also slightly better on \texttt{EC} and \texttt{IP}. The weak spots are equally informative. \texttt{PLS} drops well below the legacy rule system, and \texttt{AG} remains poor for both systems. These results suggest that the next round of work should focus on grading prompts and personalized-learning templates rather than further broadening the routing stack.

\FloatBarrier

\subsection{Ablation Snapshot}
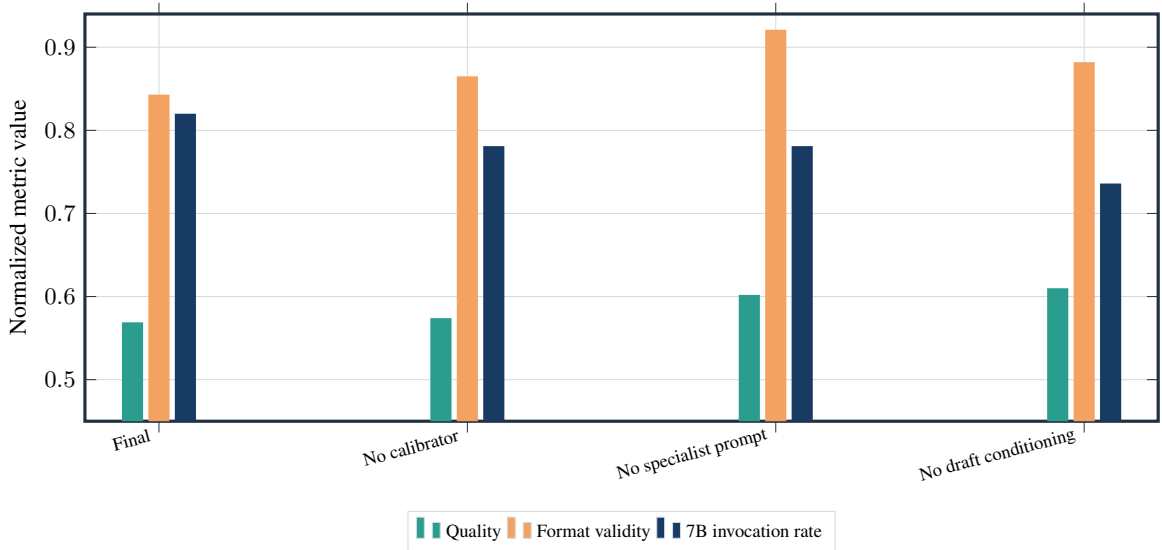
\begin{figure}[H]
  \centering
  \begin{tikzpicture}
\begin{axis}[
  width=0.96\linewidth,
  height=7.0cm,
  ybar,
  ymin=0.45,
  ymax=0.94,
  symbolic x coords={Final,No calibrator,No specialist prompt,No draft conditioning},
  xtick=data,
  x tick label style={rotate=15, anchor=east, font=\scriptsize},
  ylabel={Normalized metric value},
  axis line style={draw=paceInk, very thick},
  tick style={draw=paceInk},
  tick label style={font=\small},
  label style={font=\small},
  grid=major,
  major grid style={draw=paceGray},
  enlarge x limits=0.08,
  bar width=8pt,
  legend style={draw=paceGray, fill=white, font=\scriptsize, at={(0.5,-0.22)}, anchor=north, legend columns=3},
  clip=false
]
\addplot[fill=paceTeal, draw=none] coordinates {
  (Final,0.569)
  (No calibrator,0.574)
  (No specialist prompt,0.602)
  (No draft conditioning,0.610)
};
\addplot[fill=paceOrange, draw=none] coordinates {
  (Final,0.843)
  (No calibrator,0.865)
  (No specialist prompt,0.921)
  (No draft conditioning,0.882)
};
\addplot[fill=paceNavy, draw=none] coordinates {
  (Final,0.820)
  (No calibrator,0.781)
  (No specialist prompt,0.781)
  (No draft conditioning,0.736)
};
\legend{Quality,Format validity,7B invocation rate}
\end{axis}
\end{tikzpicture}
  \caption{Ablation snapshot on the preserved Chinese subset. Quality, format validity, and 7B invocation rate are all reported on the same \(0\)-to-\(1\) scale to highlight trade-offs among routing, prompting, and calibration choices.}
  \label{fig:ablation_metrics}
\end{figure}

The archived ablation subset adds another view of system diversity. Figure \ref{fig:ablation_metrics} compares the final cascade against three stripped variants on a \(178\)-sample Chinese subset using the saved summary artifacts. The picture is more informative than the earlier draft suggested. Removing draft conditioning raises subset quality from \(0.569\) to \(0.610\), while removing the specialist-side prompt yields the best format validity (\(0.921\)) and the lowest latency (\(23.48\)s). In other words, the current \texttt{cascade\_final} configuration is functional but not strictly dominant on this diagnostic subset.

To test whether those ablation trends transfer beyond the preserved subset, we also ran matched post-archive pilots on a separate deterministic \(76\)-sample Chinese test slice. Those pilots do \emph{not} overturn the paper's main quality ranking. \texttt{cascade\_final} remains the best quality-first configuration at \(0.553\), while the lighter variants mainly trade quality for lower latency and lower cost proxy; Appendix Table \ref{tab:live_pilot_variants} gives the exact comparison. We therefore treat the ablation as evidence of optimization headroom rather than as proof that the paper's main system should be replaced.

\FloatBarrier

\section{External EduBench Baselines and GPT-5.4 Status}
The repository also contains the official EduBench sampled baseline scores in \path{EduBench/data/all_data/model_eval_score/model_sampled_eval_scores.csv}. We aggregated those rows into paper-ready tables to retain external context. Under the existing \texttt{gpt-4o} evaluator, the strongest available sampled baseline is DeepSeek R1 with an average score of \(9.05\), followed by Qwen Max (\(9.00\)) and DeepSeek V3 (\(8.99\)); Table \ref{tab:edubench_reference} lists representative scenario-level scores.

These reference numbers are useful, but they are \emph{not directly comparable} to the deterministic \(0\)-to-\(1\) quality metric used in our current paper tables. They are sampled-response judge scores on a \(1\)-to-\(10\) scale. The correct way to align them is to re-judge the same sampled baseline responses with the desired evaluator and aggregate them into the same CSV shape.

That path is implemented locally, but as of \textbf{March 16, 2026} it remains blocked by provider configuration. The current default \texttt{GPT5\_API\_BASE} points to \url{https://api.aicodemirror.com/v1}, and probe requests to \texttt{/models}, \texttt{/chat/completions}, and \texttt{/responses} return \texttt{404}. The current secret also has an Anthropic-style prefix (\texttt{sk-ant-}), which does not match the OpenAI-compatible interface expected by the re-judge code. The missing GPT-5.4 baseline table is therefore an infrastructure issue, not an evaluation-code issue.

\begin{table}[H]
\centering
\caption{Official EduBench sampled baseline reference under the GPT-4o evaluator. These values come from the provided baseline CSV and are shown as external context only.}
\label{tab:edubench_reference}
\small
\tighttable{\begin{tabular}{lrrrrrr}
\toprule
Model & Avg. & Q\&A & IP & AG & QG & PCC \\
\midrule
\rowcolor{paceTealLight}
DeepSeek R1 & \textbf{9.05} & 9.32 & 8.78 & 8.51 & 8.98 & \textbf{9.08} \\
Qwen Max & 9.00 & \textbf{9.50} & 8.69 & \textbf{8.70} & 8.92 & 9.05 \\
DeepSeek V3 & 8.99 & 9.22 & 8.77 & 8.54 & \textbf{9.00} & 8.95 \\
Qwen2.5-14B-Instruct & 8.87 & 9.34 & 8.51 & 8.11 & 8.77 & 8.82 \\
Qwen2.5-7B-Instruct & 8.86 & 9.22 & \textbf{8.84} & 8.04 & 8.62 & 8.94 \\
\bottomrule
\end{tabular}
}
\end{table}
\FloatBarrier

\section{Limitations and Conclusion}
This paper revision is deliberately narrower than the original draft. It reports what the archived artifacts actually support:
\begin{itemize}
    \item The full main table is Chinese only because that is the only split with full archived \texttt{cascade\_final} and \texttt{rule\_v2} predictions after correction.
    \item The \texttt{1B}-only and \texttt{7B}-only baselines are diagnostic rather than archival full-benchmark runs.
    \item Routing selectivity is real, but it has not yet translated into wall-clock gains in the detached deployment used for the experiments.
    \item External sampled-baseline alignment with GPT-5.4 remains pending a valid provider configuration.
\end{itemize}

Even with these limits, the paper now tells a coherent and defensible story. Pangu-ACE is a real educational cascade, not a placeholder architecture. Its saved artifacts support a clear conclusion: a calibrated \texttt{1B} $\rightarrow$ \texttt{7B} pipeline improves output quality and output validity over the legacy rule system while showing meaningful task-dependent routing behavior. The next engineering milestone is not another model role; it is to make the current routing savings visible in wall-clock performance and to finish the GPT-5.4 baseline alignment once the provider configuration is repaired.

\appendix
\clearpage
\section{Supplementary Tables}
We include the preserved fast diagnostic and ablation tables for completeness. The ablation subset is small (\(178\) Chinese samples), so we treat it as directional evidence rather than a final ranking of components.

\begin{table}[H]
\centering
\caption{Chinese fast diagnostic subset.}
\small
\tighttable{\begin{tabular}{lrrrrrr}
\toprule
System & Samples & Quality & Format & Latency (s) & 7B rate & 1B accept \\
\midrule
\texttt{1b\_only} & 354 & 0.167 & 0.220 & 7.77 & 0.0\% & 100.0\% \\
\texttt{rule\_v2} & 354 & 0.456 & 0.686 & 17.87 & 100.0\% & -- \\
\rowcolor{paceTealLight}
\texttt{cascade\_final} & 354 & 0.510 & 0.833 & 28.20 & \textbf{77.1\%} & \textbf{22.9\%} \\
\texttt{7b\_only} & 354 & \textbf{0.591} & \textbf{0.847} & 22.32 & 100.0\% & -- \\
\bottomrule
\end{tabular}
}
\end{table}

\begin{table}[H]
\centering
\caption{English fast diagnostic subset.}
\small
\tighttable{\begin{tabular}{lrrrrrr}
\toprule
System & Samples & Quality & Format & Latency (s) & 7B rate & 1B accept \\
\midrule
\texttt{1b\_only} & 368 & 0.195 & 0.236 & 8.15 & 0.0\% & 100.0\% \\
\rowcolor{paceTealLight}
\texttt{cascade\_final} & 368 & 0.472 & 0.750 & 30.32 & \textbf{85.3\%} & \textbf{14.7\%} \\
\texttt{7b\_only} & 368 & \textbf{0.534} & \textbf{0.785} & 25.68 & 100.0\% & -- \\
\bottomrule
\end{tabular}
}
\end{table}

\begin{table}[H]
\centering
\caption{Chinese ablation subset (\(178\) samples).}
\small
\tighttable{\begin{tabular}{lrrrrr}
\toprule
Variant & Samples & Quality & Format & Latency (s) & 7B rate \\
\midrule
\rowcolor{paceTealLight}
\texttt{cascade\_final} & 178 & 0.569 & 0.843 & 27.68 & 82.0\% \\
w/o calibrator & 178 & 0.574 & 0.865 & 27.08 & 78.1\% \\
w/o specialist prompt & 178 & 0.602 & \textbf{0.921} & \textbf{23.48} & 78.1\% \\
w/o draft conditioning & 178 & \textbf{0.610} & 0.882 & 26.80 & \textbf{73.6\%} \\
\bottomrule
\end{tabular}
}
\end{table}

\begin{table}[H]
\centering
\caption{Post-archive live pilot on a matched \(76\)-sample Chinese test slice. These runs are diagnostic only, but they verify that the lighter variants mainly trade quality for efficiency rather than dominating \texttt{cascade\_final}.}
\label{tab:live_pilot_variants}
\small
\tighttable{\begin{tabular}{lrrrrrr}
\toprule
Variant & Samples & Quality & Format & Latency (s) & 7B rate & Cost proxy \\
\midrule
\rowcolor{paceTealLight}
\texttt{cascade\_final} & 76 & \textbf{0.553} & \textbf{0.895} & 25.77 & 84.2\% & 12633 \\
\texttt{cascade\_tuned} & 76 & 0.521 & 0.816 & 27.07 & \textbf{81.6\%} & 14022 \\
w/o specialist prompt & 76 & 0.493 & \textbf{0.895} & \textbf{20.38} & 80.3\% & \textbf{8732} \\
w/o draft conditioning & 76 & 0.526 & 0.868 & 23.95 & 82.9\% & 9668 \\
\bottomrule
\end{tabular}
}
\end{table}

\bibliographystyle{abbrv}
\bibliography{references}

\end{document}